\documentclass[letterpaper]{article} 
\usepackage{xcolor}
\usepackage{aaai25}  
\usepackage{times}  
\usepackage{helvet}  
\usepackage{courier}  
\usepackage[hyphens]{url}  
\usepackage{graphicx} 

\usepackage{natbib}  
\usepackage{booktabs}
\usepackage{caption} 
\usepackage{subcaption}
\usepackage{comment}
\usepackage{rotating}
\usepackage{adjustbox}
\usepackage{algorithm}
\usepackage{algorithmic}

\usepackage{newfloat}
\usepackage{listings}
\DeclareCaptionStyle{ruled}{labelfont=normalfont,labelsep=colon,strut=off} 
\floatstyle{ruled}
\newfloat{listing}{tb}{lst}{}
\floatname{listing}{Listing}
\newcommand{\answerTODO}[1]{\textcolor{red}{#1}}

\usepackage{tabularx}

\newcommand{\ignore}[1]{}
\usepackage[show]{notes}
\newcommand{\com}[1]{}

\usepackage{csquotes}

\newcommand{\revision}[1]{#1}

\title{The Value of Nothing\footnote{``Nowadays people know the price of everything and the value of nothing,'' The Picture of Dorian Gray \cite{wilde2008picture}}: \\Multimodal Extraction of Human Values Expressed by TikTok Influencers }

\makeatletter
\gdef\showauthors@on{T} 
\makeatother

\author{
    Authors
     Alina Starovolsky-Shitrit\equalcontrib\textsuperscript{\rm 1},
     Alon Neduva\equalcontrib\textsuperscript{\rm 2},
     Naama Appel Doron\textsuperscript{\rm 1},
     Ella Daniel\textsuperscript{\rm 1},
     Itamar Gafni\textsuperscript{\rm 1},
     Oren Tsur\textsuperscript{\rm 2}   
}
\affiliations{
    \textsuperscript{\rm 1}Tel-Aviv University\\
    \textsuperscript{\rm 2}Ben-Gurion University of the Negev\\
    alinashitrit@mail.tau.ac.il,
    neduva@post.bgu.ac.il, \{naamaapp, della\}@tauex.tau.ac.il, orentsur@bgu.ac.il\\
    
}

\begin{document}

\maketitle

\begin{abstract}
Societal and personal values are transmitted to younger generations through interaction and exposure. Traditionally, children and adolescents learned values from parents, educators, or peers. Nowadays, social platforms serve as a significant channel through which youth (and adults) consume information, as the main medium of entertainment, and possibly the medium through which they learn different values. In this paper we extract implicit values from TikTok movies uploaded by online influencers targeting children and adolescents. We curated a dataset of hundreds of TikTok movies and annotated them according to the well established Schwartz Theory of Personal Values. We then experimented with an array of language models, investigating their utility in value identification. Specifically, we considered two pipelines: direct extraction of values from video and a 2-step approach in which videos are first converted to elaborated scripts and values are extracted from the textual scripts.

We find that the 2-step approach performs significantly better than the direct approach and that using a few-shot application of a Large Language Model in both stages outperformed the use of a fine-tuned Masked Language Model in the second stage. We further discuss the impact of continuous pretraining and fine-tuning and compare the performance of the different models on identification of values endorsed or confronted in the TikTok. Finally, we share the first values-annotated dataset of TikTok videos.

To the best of our knowledge, this is the first attempt to extract values from TikTok specifically, and visual social media in general. Our results pave the way to future research on value transmission in video-based social platforms.
\end{abstract}

%

\section{Introduction}
\label{sec:intro}


The transmission of values between generations is a key process for the preservation of society \cite{schonpflug2001intergenerational}. Classic research assumed that transmission of values is carried out mainly by parents \cite{Grusec2011} or schools \cite{Berson2016}. However, youth spend hours every day on social media \cite{Smahel2020}, where they find new role models \cite{Tolbert2019}. The social media environment has been shown to strongly influence children's behavior, in the context of consumer behavior \cite{Coates2019}. However, influencers on social media promote more than products: they form communities and promote shared values \cite{himelboim2023social}.

The Schwartz Theory of Personal Values \cite{Schwartz1992} offers a well-studied framework for the investigation of human motivation. It defines values as abstract goals, describing the end states people aspire to achieve in their lives, such as success, independence, and care.
These goals cover a wide range of motivations, with different individuals and societies varying in the prioritization they place on different values \cite{Schwartz1992, Sagiv2022}. Values are expressed through a wide variety of judgments, evaluations, and behaviors \cite{Sagiv2021}. The theory identifies 19 core values, characterized by the boader goals they convey or serve \cite{schwartz2012overview}. Personal values, as defined by Schwartz (see Table \ref{tab:values}), were consistently identified in different contexts, such as national cultures throughout the world \cite{Sagiv2022}, organizational cultures \cite{Daniel2024}, traditional media \cite{Kiesel2024}, and online discourse \cite{vanDerMeer2023differences}.

In this work, we suggest that social media influencers targeting youth, can be characterized not only by genre, topic, and style, but also by the values they endorse and the values they confront and disregard. To list a number of examples, an influencer hosting a competition between followers endorses values of achievement; an influencer presenting a unique craft project endorses values of creativity; influencers who pull pranks on strangers confront the values of respect for social norms. A number of values can be endorsed/confronted in the same piece of content. For example, in one video (script available in Table \ref{tab:tennis_script}), a gaming influencer fails to return tennis balls served by a professional tennis player. The value of \texttt{Achievement} (trying to properly return fast serves) is endored in the video. At the same time,  the value of \texttt{Face} is confronted, as the influencer is humiliated (shows himself failing again and again to no avail).

How can we identify the values transmitted to youth through social media? Some recent studies identified values prevalent in written social media using a variety of methods e.g. \cite{Kumar2018_MLAlgorithms,silva2020} ranging from linguistic inquiry and word count (LIWC), word matching to masked language models, such as BERT. However, current trends in social media lean heavily toward visual based media, such as TikTok \cite{Smahel2020, rejeb2024mapping}. The shift toward visual mediums introduces a further challenge, since it requires analyzing a clip in a multimodal way, combining text, audio, and video. In addition, the clips are typically very short, providing rich content through limited stimuli. To the best of our knowledge, this is the first attempt to extract values from TikTok specifically, and visual social media in general. 

In this work we have curated and annotated a unique corpus of 885 TikToks, posted by influencers targeting children and adolescents, covering a variety of genres prevalent in TikTok (see Section \ref{sec:data} for details). The corpus is useful for the investigation of values prevalent in this platform, and will be open for analysis. We experiment with and compare an array of algorithms of value classification, including fine-tuned Masked Language Models and Large Language Models. 
Specifically, we consider two main pipelines: a direct extraction from LLMs, leveraging their multimodal capabilities, and a 2-step extraction approach in which a video is first converted to a detailed textual script (master scene heading, action, dialogue) and the values are extracted from the text, rather than directly from the video. In spite of the multimodal complexity, we achieve a high F-score, higher than the state-of-the-art results reported on the task of extracting values from texts \cite{schroter2023adam, legkas2024hierocles}.

We find that the 2-step approach performs better than the direct approach and that using a few-shot application of a Large Language Model outperforms a trainable Masked Language Model in the second step. We further discuss the impact of fine-tuning and compare the performance of the different models on the endorsed  and the confronted values.

\begingroup
\renewcommand{\arraystretch}{1.5} 

\begin{table*}[ht!]
    \centering
    \small
    \begin{tabular}{ll}
    Value & ~~~~~~~~~~~~~~~~~~~~~~~~~~~~~~~~~~~~~~Definition \\ [2pt]
    \hline
    SELF-DIRECTION–THOUGHT & Freedom to cultivate one’s own ideas and abilities \\ \hline
SELF-DIRECTION–ACTION &  Freedom to determine one’s own actions \\ \hline
STIMULATION  & Excitement, novelty, and change \\ \hline
HEDONISM  & Pleasure and sensuous gratification \\ \hline
ACHIEVEMENT  & Success according to social standards \\ \hline
POWER–DOMINANCE & Power through exercising control over people \\ \hline
POWER–RESOURCES & Power through control of material and social resources \\ \hline
FACE  & Security and power through maintaining one’s public image and avoiding humiliation \\ \hline
SECURITY–PERSONAL   & Safety in one’s immediate environment \\ \hline
SECURITY–SOCIETAL  & Safety and stability in the wider society \\ \hline
TRADITION  & Maintaining and preserving cultural, family, or religious traditions \\ \hline
CONFORMITY–RULES  & Compliance with rules, laws, and formal obligations \\ \hline
CONFORMITY–INTERPERSONAL  & Avoidance of upsetting or harming other people \\ \hline
HUMILITY  & Recognizing one’s insignificance in the larger scheme of things \\ \hline
BENEVOLENCE–DEPENDABILITY  & Being a reliable and trustworthy member of the in-group \\ \hline
BENEVOLENCE–CARING  & Devotion to the welfare of in-group members \\ \hline
UNIVERSALISM–CONCERN  & Commitment to equality, justice, and protection for all people \\ \hline
UNIVERSALISM–NATURE  & Preservation of the natural environment \\ \hline
UNIVERSALISM–TOLERANCE  & Acceptance and understanding of those who are different from oneself \\ \hline
    \end{tabular}
    \caption{Value definitions, Schwartz Theory of Personal Values \cite{Schwartz1992,schwartz2012overview}}
    \label{tab:values}
\end{table*}

\endgroup

\section{Related Work}
\label{sec:related}

\paragraph{The TikTok Influencers Ecosystem} 
Youth spend hours on various social media sites \cite{Smahel2020}, with TikTok gaining increased popularity \cite{anderson2023teens, woodward2024}. TikTok is characterized by engaging short videos that can include widespread use of filters, and background music \cite{rejeb2024mapping}. The use of a custom ``For You'' page and ``Like'' button increases immersion in the application \cite{montag2021psychology}. 

Social media in general, and TikTok specifically, are characterized by a large number of Social Media Influencers (SMIs). Social media influencers are social media users who gained a large following \cite{ki2020influencer} by creating and curating content that portrays an appealing online personality \cite{tafesse2021followers}. Influencers form parasocial relationships with their followers: members of the audience feel close to influencers, as if they know them intimately \cite{reinikainen2020you}. Influencers are especially likely to form such  relationships because of social media's interactive nature \cite{conde2023micro, lou2019influencer}, and the disclosure of the influencers' personal lives \cite{conde2023micro}. Followers often identify with their favorite influencers \cite{lou2019influencer}, making them likely role models for young viewers.

Ample research supports the understanding that children and adolescents can learn from mass traditional media \cite{prot2015media}, as well as social media \cite{craig2021prosocial}. From the media, youth can acquire information \cite{dorey2009role}, but also form their worldviews \cite{duffett2017influence}. A recent study established that the following of social media influencers is associated with higher importance ascribed to materialistic values among Finish youth \cite{tuominen2023modern}. We suggest that in order to further investigate the role of influencers in youth value transmission, we must identify the values they embrace. 

\paragraph{Personal Values}
Values are abstract goals describing end-states individuals aspire to achieve in their lives \cite{Sagiv2022,Schwartz1992}. Values direct a wide variety of behaviors, ranging between life-altering choices such as career preferences \cite{Lipshits2024}, and daily behaviors, such as the choice of touristic destinations \cite{Beena2015}. 

The Schwartz Theory of Personal Values is the leading theory of values in social psychology, well-validated in hundreds of samples across the world \cite{Sagiv2017}. Importantly, the theory extends beyond moral values, such as justice and care, to define a variety of motivations that can drive individuals \cite{Twito2022}, including, for example, safety, excitement, and influence. Moreover, the values are organized in a dynamic system of inter-relations, as the pursuit of one value may contrast the pursuit of conflicting values, and enhance the pursuit of compatible ones \cite{Schwartz1992}. These inherent trade-offs require individuals to prioritize among values. The inter-relations can be summarized into two main axes. One describes the tension between self-transcendence (care for others, close and distant) and self-enhancement (focus on advancement of self-interest). The second describes the tension between conservation (preservation of the status-quo) and openness to change (embracing new experiences). The list of Schwartz values and their short description is available at Table \ref{tab:values}.

Research established that children and adolescents hold meaningful values, following the content and structure defined by the Schwartz Theory of Personal Values \cite{DanielMisgav2024, Daniel2019, Knafo2023}. It is often agreed that children's values are mainly learned from the social environment, in a process of value transmission \cite{Twito2020}. However, the social environment is varied and includes many socialization agents. Such agents are parents \cite{Grusec2011, KnafoBarni2020}, schools \cite{Berson2016}, the peer group \cite{BenishOreg2022,BenishPeers2024}, and the national culture \cite{Daniel2012}. As social media becomes an important factor in the lives of adolescents, it is crucial to learn whether individuals within this sphere, such as influencers, become socialization agents teaching personal values to youth. 

\paragraph{Values Social Media}
Schwartz values were identified in content displayed across social media platforms, such as Instagram \cite{trillo2021does}, Reddit \cite{chen2014understanding}, Facebook and Twitter \cite{Kumar2018_MLAlgorithms}. Values were presented in social media content in varied ways, such as memes \cite{shifman2019internet}, and new-years’ resolutions \cite{hallinan2023value}. However, little is known about the prevalence of Schwartz values in youth-directed media. Some research exists on the prevalence of values in child-directed mass traditional media. An analysis of television-shows addressing 9-11 year olds, showed high representation of fame values \cite{uhls2011rise}. Similarly, television shows addressing young children, displayed work as an expression of a large variety of values, including benevolence, security, conformity, achievement, universalism, stimulation and hedonism \cite{aharoni2024integrating}. Our work aims at opening the door to extract values at scale by developing an AI-based value detection in TikTok movies. 

\paragraph{AI-Based Value Classification} The field of automated value detection has evolved significantly. Focusing on methods for classifying values according to Schwartz's Theory of Personal Values \cite{schwartz2012overview}, early approaches leveraged the Linguistic Inquiry and Word Count (LIWC) \cite{LIWCpennebaker2015development}, or a targeted dictionary \cite{ponizovskiy2020development} detecting values through lexical matching. Values were detected in texts produced across various platforms, including Twitter \cite{kadic2024no}, narrative texts \cite{fischer2022opportunities}, song lyrics \cite{demetriou2024towards}, and social media platforms such as Reddit \cite{oulahbib2024measuring} and Facebook \cite{Kumar2018_MLAlgorithms}. Notable advances include the development of stack models combining LIWC features trained on Facebook data with cross-platform validation on Twitter \cite{silva2020}. VALUENET \cite{qiu2022valuenet}, introduced a large-scale text dataset and implemented a Transformer-based value regression model for human value modeling in dialogue systems. 

Recent developments in value prediction were introduced with the application of masked language models. The SemEval-2022 \cite{kiesel2022identifying} and CLEF-2024 \cite{kiesel2024overview} competitions  focused on value detection in argumentative texts and multi-lingual traditional media contexts respectively. CLEF-2024 was the first to identify both endorsed and confronted values in texts. These competitions demonstrated the effectiveness of BERT-based architectures \cite{schroter2023adam, legkas2024hierocles} in capturing nuanced value expressions across two different text genres and languages. It also demonstrated that algorithms developed in one context may not succeed in identification of values in a another context \cite{kiesel2024overview}, asserting the importance of contextual models in value detection. 

Generative AI methods have not yet been utilized for value detection, with the exception of the FULCRA project \cite{yao-etal-2024-value}, leveraging LLM-generated text pairs for safety-oriented value detection. Our work extends these methodological advances by pioneering value detection in TikTok's unique multimodal environment. Unlike previous approaches that focused on single modalities, TikTok content presents a complex interplay of written text, audio, and visual elements. To address this challenge, we conduct a comparative analysis of two leading approaches: BERT-based architectures and LLM-based methods. By evaluating these distinct methodologies on the same TikTok dataset, we provide the first systematic comparison of their effectiveness in detecting values within short-form video content. This novel application to multimodal social media content, combined with our comparative analysis, can contribute valuable insights to both social media analysis and automated value detection research, particularly for platforms where meaning is conveyed through multiple channels simultaneously.


\begin{figure}
    \centering
    \includegraphics[width=1\linewidth]{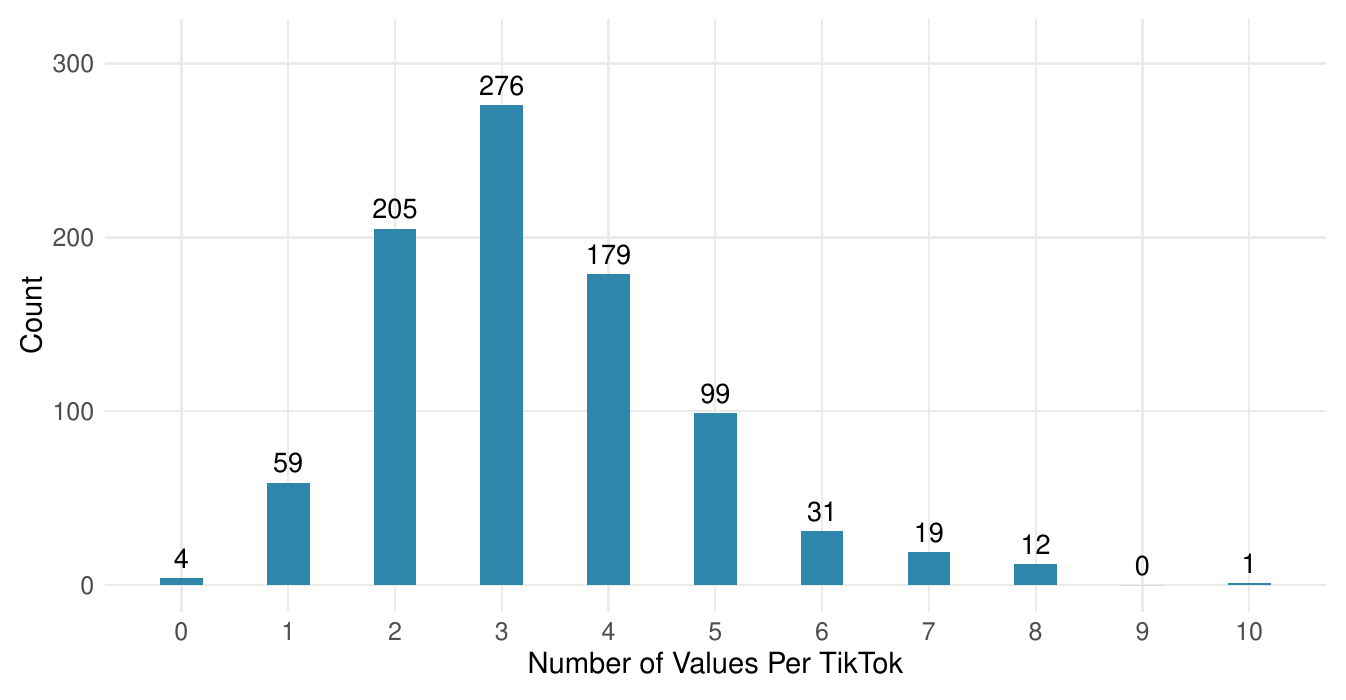}
    \caption{Unique values per TikTok}
    \label{fig:hist_values_per_video}
\end{figure}

\begin{figure*}
    \centering
    \includegraphics[width=1\linewidth]{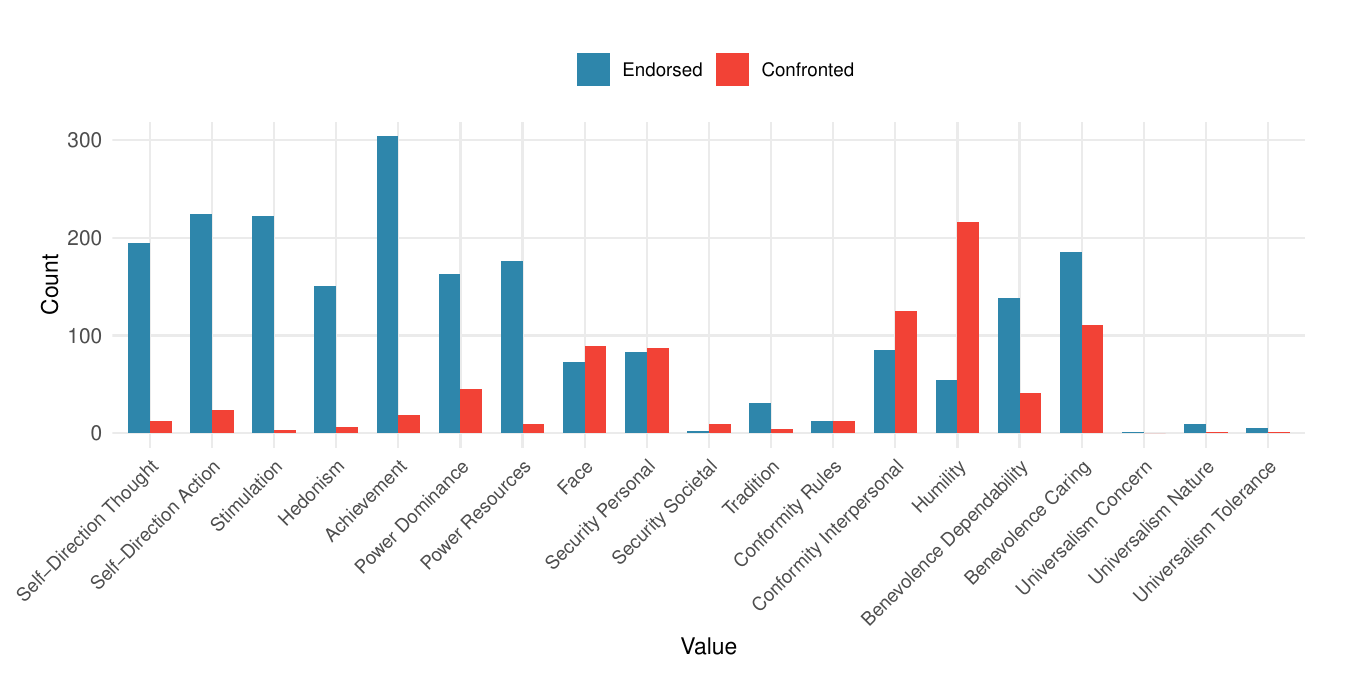}
    \caption{Occurrences of each value in the annotated dataset. Blue bars indicate the appearance of an endorsed value (e.g., \emph{achievement}) and red bars indicate the confronted value (e.g., \emph{apathy} and \emph{lack of ambition}, conflicting with \emph{achievement}).}
    \label{fig:value_counts}
\end{figure*}

\section{Data and Annotation}
\label{sec:data}

\subsection{Data Curation}
We identified relevant TikTok influencers using anonymous surveys, asking children and adolescents (local IRB approved) to list the TikTok content creators they follow. We also prompted three language models (ChatGPT, Gemini, Bing) to provide lists of prominent influencers with clout over young users. Two RAs verified the accounts and confirmed they indeed target children or adolescents
and classified them by genre. This process resulted in a list of 91 influencers, active in the following genres: beauty and skincare, lifestyle, entertainment and pranks, crafts and DYI, vlogging, gaming and dancing, singing and lip-syncing.  
We sampled ten TikToks from the page of each influencer, including all TikToks pinned by the influencers, if existed.
The full dataset contains $N = 885$ movies. 

The crawling and curation were done with slight modifications to the Teather API \cite{Teather_TikTokAPI_2024}, collecting only publicly available data. Such data curation, handling, processing and analysis, requires careful consideration of ethical aspects and potential harms, from privacy issues to intellectual property rights. We discuss potential harms, ethical considerations and handling protocols in Section \ref{sec:ethics}.

\subsection{Data Annotation}
Annotation guidelines were based on the Schwartz theory of personal values \cite{Schwartz1992,schwartz2012overview}. The guidelines provide a definition for each of the 19 values, along with examples of how values are expressed. For each TikTok and value, annotators were asked to mark whether the values is endorsed (expressed or pursued in the TikTok; 1),  absent (not referenced explicitly or implicitly; 0) or confronted (contradicted or hindered; -1). Each TikTok was independently annotated by two annotators from a team trained by the authors. A third annotator consolidated the annotations by resolving discrepancies. The annotation guidelines are available in Appendix \ref{app:subsec:values_prompt}. We converted each TikTok annotation to a vector of 38 binary labels representing the 19 endorsed values and 19 confronted values.

The annotated dataset of 885 TikToks contains 2933 (endorsed/confronted) values. The distribution of values per TikTok is presented in Figure \ref{fig:hist_values_per_video} and the frequency of each value in the dataset is presented in Figure \ref{fig:value_counts}. 
The annotated dataset (TikTok IDs and annotations) is available at [URL].
We used Gwet's AC1 coefficient to assess inter-rater agreement. This coefficient estimates reliability beyond 2 rateres, and has been demonstrated robust against prevalence imbalance and marginal probability issues \cite{wongpakaran2013comparison}. The AC1 agreement was 0.77, indicating substantial inter-annotator agreement.


\section{Computational Approach}
\label{sec:computational} 

We explore a number of computational approaches, differing in two fundamental ways: (i) A direct approach vs. a two-step approach, and (ii) Supervised classification vs. unsupervised classification using a LLM. 

\paragraph{Direct vs. 2-step Extraction} We use an LLM as a classifier in two settings: direct and indirect (2-steps). In the \emph{direct setting}, we extract the values directly from the TikTok \texttt{mp4} video file, either using an LLM, prompting it with the task definition in a few-shots manner, \revision{or in a supervised manner, after fusing the video, audio and textual representations ({\bf MLMs vs. LLMs} and  {\bf Multimodality with Masked Video Autoencoder} below)}. 

The \emph{two-step} approach allows us to divide and conquer: we first convert a TikTok clip to a full \emph{textual} script containing master scene heading, action, and dialogue, then extract the values from the textual script.

The textual scripts are generated by a dedicated prompt. This prompt was decided after iterative prompt engineering on a representative subsample of 18 videos. Our prompt design prioritized capturing three key video elements: visual content, verbal narration, and text captions. Two of the authors independently evaluated script quality, ensuring the conversion accurately captures all three elements. The conversion was performed with Gemini 1.5 Pro. 
An example of a generated script is presented in Table \ref{tab:tennis_script}. The prompt we use to convert clips to scripts is available in Appendix \ref{app:subsec:script_prompt}.

\begin{figure}[h!]
    \centering
    \includegraphics[width=1\linewidth]{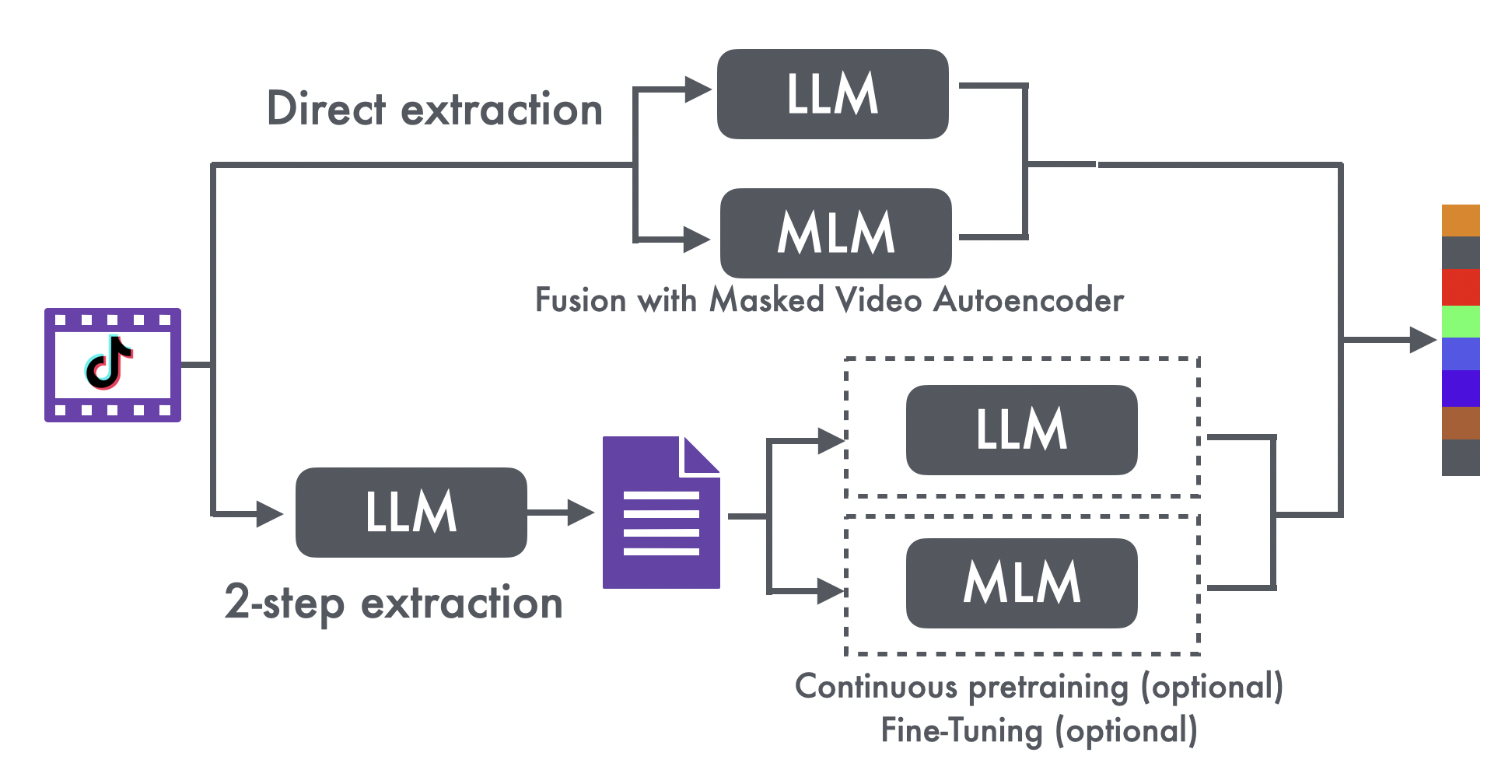}
    \caption{Direct and indirect extraction of values from TikTok movies. continuous pre-training the textual models on scripts is optional. LLMs are used in a zero/few-shot manner. The MLM must be trained on a subsample in a supervised manner.}
    \label{fig:architecture}
\end{figure}

Using the generated script, the extraction on the second 2-step could be done either in a supervised manner, including continuous pretraining and fine-tuning, or in a few-shot manner, using the exact same prompt used in the direct approach, albeit, replacing the word `movie' with `script'. The two pipelines are illustrated in Figure \ref{fig:architecture}.

\paragraph{MLMs vs. LLMs} State-of-the-Art models such as ChatGPT-4o \cite{openai2024gpt4} and Gemini 1.5 Pro \cite{geminiteam2024gemini} are multimodal, processing different modalities such as text, sound and video. Using them allows for a direct approach, analyzing all aspects of the TikTok simultaneously. Moreover, the LLMs allow for a few-shot extraction, with no need for supervised training. Using an LLM, either in the direct or the 2-step pipeline, the extraction is done by prompting the model with the annotation guidelines, alongside a request to identify the values in a given TikTok (either directly from the \texttt{mp4} file or from the generated script). All prompts are available in Appendix \ref{app:prompts}.
The LLMs used were Gemini 1.5 Pro and Mistral 7B \cite{jiang2023mistral}. 

On the other hand, the use of masked models, either MLMs like BERT or video autoencoders, allows model training, given that annotated data  is available. The supervised approach is argued to better suit tasks that require nuanced understanding of the content \cite{ziems24llms4css}. Masked models can be used in both the direct and the two-step pipelines. 
In this work we experiment with roBERTa-large \cite{liu2019roberta}, DeBERTa-V2-xlarge \cite{hedebertav3_2023}, \revision{and the Longformer \cite{Beltagy2020Longformer}}.

\revision{\paragraph{Masked Language Models} We use pre-trained transformer models with a binary classification head. We fine-tuned the classification head with 3-fold cross-validation with early stopping to prevent overfitting. We selected the best-performing epoch for each fold based on the F1 score, and then averaged these optimal results across all folds to obtain final performance metrics (commonly referred to as "best-epoch averaging", recommended by \cite{reimers2017reporting}.) }


\begin{table}[ht!]
\centering
\scriptsize
\begin{tabularx}{\linewidth}{>{\raggedright\arraybackslash}p{8cm}}
\hline
\texttt{Genre: Sports/Challenge}\\
\texttt{Sound: Yes}\\
\texttt{NARRATOR: "This pro tennis player Taylor Fritz with one of the fastest serves on tour. His fastest serve ever was 240 KM/H, but can I, a person who's never played tennis, return one?"}\\
\texttt{[pro]}\\
\texttt{TAYLOR FRITZ is shown on a tennis court serving, then staring into the camera.}\\
\texttt{NARRATOR: \#9 IN THE WORLD with one fastest serves 240KM/H I never return}\\
\texttt{The NARRATOR is shown holding a tennis racket, about to return a serve.}\\
\texttt{[SERVE 1]}\\
\texttt{TAYLOR FRITZ serves the ball. The NARRATOR misses.}\\
\texttt{NARRATOR: "Oof"}\\
\texttt{[SERVE 4]}\\
\texttt{TAYLOR FRITZ serves. The NARRATOR misses again.}\\
\texttt{NARRATOR: "Oh, yo, that was..."}\\
\texttt{[SERVE 10]}\\
\texttt{TAYLOR FRITZ serves. The NARRATOR misses.}\\
\texttt{NARRATOR: "F*ck"}\\
\texttt{[SERVE 18]}\\
\texttt{TAYLOR FRITZ serves. The NARRATOR misses.}\\
\texttt{NARRATOR: "F*ck. [giggles] Yo!"}\\
\texttt{[SERVE 24]}\\
\texttt{TAYLOR FRITZ serves. The NARRATOR misses.}\\
\texttt{NARRATOR: "F*ck, I thought I could do this."}\\
\texttt{[SERVE 32]}\\
\texttt{TAYLOR FRITZ serves. The NARRATOR misses.}\\
\texttt{NARRATOR: "F*ck no, I cannot!"}\\
\texttt{[SERVE 38]}\\
\texttt{TAYLOR FRITZ serves. The NARRATOR misses.}\\
\texttt{[SERVE 47]}\\
\texttt{TAYLOR FRITZ serves. The NARRATOR hits the ball back.}\\
\texttt{NARRATOR: "Oh! Where'd it go?"}\\
\texttt{[50TH SERVE]}\\
\texttt{TAYLOR FRITZ serves for the last time.}\\
\texttt{NARRATOR: "Oh, come on."}\\
\hline
\end{tabularx}
\caption{The script of video ID 7341895431883967746}
\label{tab:tennis_script}
\end{table}

\revision{\paragraph{Multimodality with Masked Video Autoencoder} VideoMAE \cite{tong2022videomae}, a masked autoencoder pretrained specifically on videos, is reported to achieve impressive results even over small datasets. We extract the visual embeddings using VideoMAE, audial embeddings using Wav2Vec2-Base \cite{baevski2020wav2vec} and the textual embeddings using the SentenceTransformer \cite{reimers2019sentence}. Fusion is done by concatenation. These fused embeddings were then fed into an MLP classification head.  We refer to this fused multimodel F-VideoMAE. For more details see Appendix \ref{app:subsec:videoMAE}.}

\paragraph{Evaluation Procedure and Metrics} We evaluate the different pipeline and settings using an the standard F1-score per value and over all values (macro and weighted). \revision{We also use the recall value in a BLEU-like manner. While the different variations of the F-score allows us to evaluate the performance of each system in the detection of each specific value, as well as the overall performance. This Recall/BLEU value, inspired by the BLEU \cite{papineni2002bleu} allows us to assess the performance on the \emph{movie} level, measuring how many of the values conveyed in a specific TikTok were actually recovered by the algorithm, regardless of the specific mixture of values. We report and discuss the average Recall/BLEU values.}

\revision{In the unsupervised settings (direct and the 2-stage approaches) the results are computed over the whole dataset. Settings involving supervised training were conducted in a 3-fold cross-validation manner, each fold uses two thirds of the data for training and a third for testing. Results are reported only for values appearing in at least 5\% of the movies, in order to allow enough data in the train/test split.}

\paragraph{Compute Resources} All experiments were executed on Google Colab Enterprise runtime environment with an NVIDIA Tesla A100 GPU (a2-highgpu-1g configuration).


\begin{table}[th!]
\centering
\small
\begin{tabular}{l@{}|cc|cc|cc}
     \multicolumn{1}{c}{} & \multicolumn{2}{c}{\boldmath{Endorsed}} & \multicolumn{2}{c}{\boldmath{Confronted}} & \multicolumn{2}{c}{\boldmath{All}}\\ [2pt]
     & W & M & W & M & W & M \\ [2pt]
   \hline \\[-0.8em]
   F-VideoMAE$^d$ & 0.23 & 0.20 & 0.22 & 0.17 & 0.23 & 0.19 \\
   Gemini$^d$ & 0.46 & 0.40 & 0.26 & 0.27 & 0.4 & 0.37 \\
   \hline
   RoBERTa$^{2s}_{pt}$ & 0.26 & 0.23 & 0.24 & 0.21 & 0.26 & 0.23 \\
   DeBERTa$^{2s}_{PT}$ & 0.3 & 0.27 & 0.24 & 0.22 & 0.29 & 0.25 \\
   Longformer$^{2s}$ & 0.34 & 0.30 & 0.28 & 0.20 & 0.33 & 0.27 \\
   DeBERTa$^{2s}$ & 0.36 & 0.33 & 0.35 & 0.32 & 0.36 & 0.33 \\
   Mistral$^{2s}$ & 0.39 & 0.36 & 0.26 & 0.26 & 0.36 & 0.33 \\
   RoBERTa$^{2s}$ & 0.42 & 0.38 & \textbf{0.38} & 0.35 & 0.41 & 0.37 \\
   Gemini$^{2s}$ & \textbf{0.48} & \textbf{0.43} & 0.36 & \textbf{0.37} & \textbf{0.45} & \textbf{0.41} \\
\end{tabular}
\caption{Weighted (W) and Macro (M) F-Score results for the Endorsed Values , the Confronted Values and All. The direct and two-steps settings are indicated by $d$ and $2s$ (for \emph{2-step}), respectively; $PT$ indicates continuous Pre-Training.}
\label{tab:results}
\end{table}

\begin{figure*}[ht!]
     \centering
     \begin{subfigure}[b]{0.49\textwidth}
         \centering
         \includegraphics[width=\textwidth]{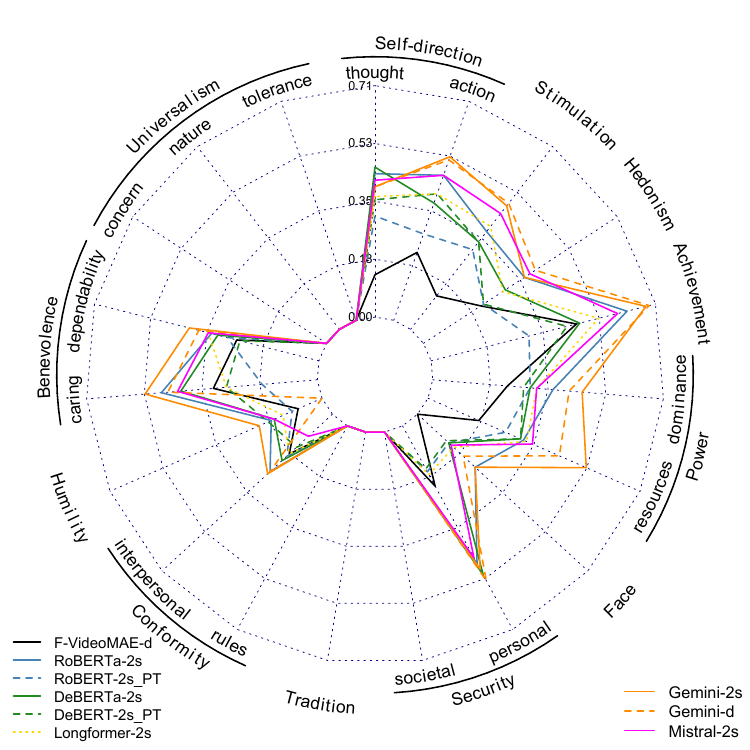}
         \caption{Endorsed Values}
         \label{subfig:radar_pos}
     \end{subfigure}
     \hfill
     \begin{subfigure}[b]{0.49\textwidth}
         \centering
         \includegraphics[width=\textwidth]{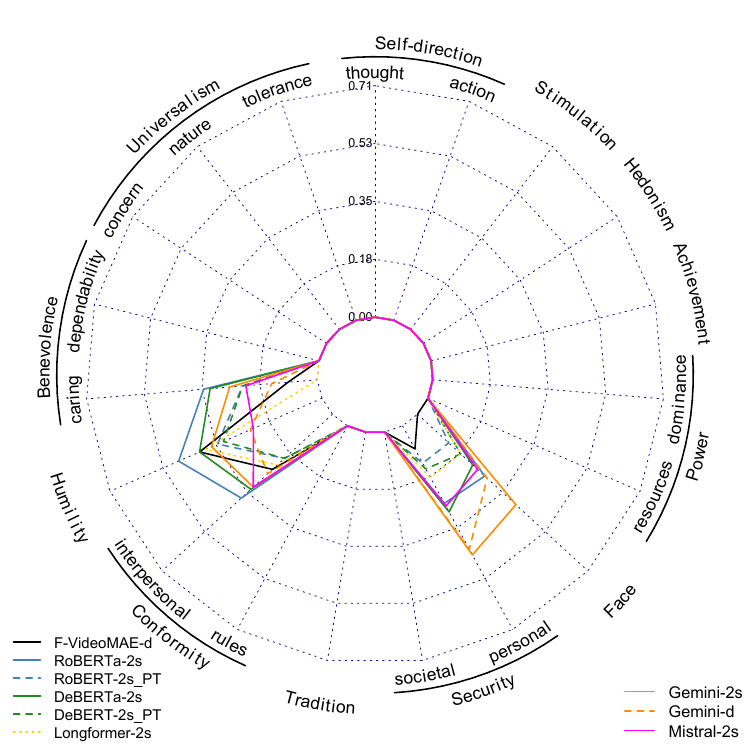}
         \caption{Confronted Values}
         \label{subfig:radar_neg}
     \end{subfigure}
    \caption{F-Scores per each model and value. Values appearing less than 5\% of TikToks were not included in the BERT based MLM models, thus excluded from the analysis (appearing with score 0 in the figure).}
    \label{fig:results_radar}
\end{figure*}

\section{Results and Analysis}
\label{sec:results}

\paragraph{Overall Results} Table \ref{tab:results} presents the weighted and the macro F-scores achieved by each of the models. Best results \emph{overall} were obtained by Gemini$^{2s}$ (using the 2-stage pipeline, prompting Gemini in both stages) with 0.45 (0.41) weighted  (macro) F-score. 
Separating \emph{endorsed} and \emph{confronted} values, this setting achieves best results (weighted and macro) on the endorsed values, and best macro results on the confronted values. While the RoBERTa$^{2s}$ achieved the highest weighted F-score on the confronted values, Gemini$^{2s}$ is a close second, with 0.36, compared to 0.38 by RoBERTa$^{2s}$. Interestingly, the direct pipeline (Gemini$^d$) tend to underperform, compared to the use of the Gemini or the RoBERTa in a 2-stage pipeline.  

\revision{These results suggest that while Gemini was trained in a multimodal manner over text and video, nuanced classification better done over texts. Indeed, the 2-stage approach  harnesses the multimodal capabilities of Gemini to convert videos to detailed textual scripts (master scene heading, action, dialogue) and then harness the capabilities of the LLM as a strong few-shot learner. The detailed script generated in the first stage helps mitigating the challenges demonstrated by \citet{ziems24llms4css}.}

\revision{\paragraph{Results per Movie (Recall/BLEU)} Focusing on the \emph{movie} level, rather than on the \emph{value} level, the 2-stage pipeline proved to achieve the best results (Table \ref{tab:bleu_scores}). This is consistent with the results presented above. Gemini$^{2s}$ demonstrates superior performance on the endorsed values and achieves the best overall results. However, when focusing specifically on confronted values at the \emph{movie} level, RoBERTa$^{2s}$ outperforms Gemini$^{2s}$. We attribute this gap to the fact that confronted values are more opaque and therefore harder to detect without explicit supervision. At the \emph{movie} level, evaluation is based on recall rather than F1; for confronted values, which are more opaque, this favors models such as RoBERTa$^{2s}$ that exhibit higher sensitivity. This increased sensitivity, however, is accompanied by a substantially higher rate of false positives, which leads to lower precision, as evident at the \emph{value} level (Table \ref{tab:Confronted1_Appendix}). However, at the \emph{movie} level, where evaluation emphasizes recall rather than precision, this reduction in precision is not directly reflected. In summary, for the confronted values, at the \emph{value} level this results in comparable F1 scores to Gemini$^{2s}$ (Table \ref{tab:results}), while at the \emph{movie} level, the higher recall of RoBERTa$^{2s}$ leads to improved performance.}  

\begin{table}
\centering
\begin{tabular}{lccc}

Model & Endorsed  & Confronted  & All Values \\
\hline
Gemini$^d$ & 0.51 & 0.2 & 0.36 \\
Mistral$^{2s}$ & 0.51 & 0.26 & 0.39 \\
roBERTa$^{2s}_{PT}$ & 0.31 & 0.30 & 0.31 \\
DeBERTa$^{2s}_{PT}$ & 0.33 & 0.30 & 0.31 \\
Gemini$^{2s}$ & {\bf 0.58} & 0.32 &{\bf 0.45} \\
DeBERTa$^{2s}$ & 0.39 & 0.39 & 0.39 \\
roBERTa$^{2s}$ & 0.43 & {\bf 0.44} &  0.43 \\
\hline
\end{tabular}
\caption{\revision{Recall/BLEU results. The direct and two-steps pipelines are indicated by $d$ and $2s$ (2-step), respectively; $PT$ indicates continuous Pre-Training.}}
\label{tab:bleu_scores}
\end{table}

\revision{\paragraph{Results per Value} The results \emph{per value} are presented in the radar plots in Figure \ref{fig:results_radar}. The highest F-score for endorsed values (0.69) was obtained for \texttt{Achievement}, followed by \texttt{Security Personal} and \texttt{Power Resources} (0.54 and 0.53, respectively) while the highest score for a confronted value (0.48) was obtained for \texttt{Humility}, followed by \texttt{Security Personal} and \texttt{Face} (0.45, 0.41). Indeed, these are three of the five most frequent confronted values, see Figure \ref{fig:value_counts}.\footnote{ We note that the numbers reported here result from different settings, see  detailed results in Appendix \ref{app:detailed_results}.}}

\revision{\paragraph{Endorsed vs. Confronted Values} Most of the endorsed values are more frequent in the dataset (Figure \ref{fig:value_counts}) and were better detected than confronted values (Figure \ref{fig:results_radar} and Table \ref{tab:results}).  All models showed a performance drop between the endorsed and confronted values. The three top performing models (pipelines), Gemini$^{d}$, RoBERTa$^{2S}$, and Gemini$^{2s}$, exhibited an interesting pattern. The performance drop for Gemini$^{d}$ (direct pipeline) was nearly twice as large as that of Gemini$^{2s}$, which itself was about three times the drop observed for RoBERTa$^{2s}$. The weighted F1-scores over the endorsed values were 0.48, 0.46, and 0.42 for Gemini$^{2s}$, Gemini$^{d}$, and RoBERTa$^{2s}$, respectively. Over the confronted values, the F1-scores decreased to 0.36 ($\Delta=0.12$), 0.26 ($\Delta=0.20$), and 0.38 ($\Delta=0.04$), respectively.

This drop in performance for confronted values is also evident at the \emph{movie} level, where we aggregate over all values per video (Table \ref{tab:bleu_scores}). Gemini$^{2s}$ maintains the best overall performance (0.45), while RoBERTa$^{2s}$ outperforms Gemini$^{2s}$ on the confronted values (0.44 vs. 0.32), reflecting its higher sensitivity in detecting these opaque manifestations. The direct pipeline Gemini$^{d}$ suffers a large drop (0.20), consistent with the pattern observed at the \emph{value} level (Table \ref{tab:results}).

It seems the manifestations of confronted values pose a challenge even for trained annotators: their detection requires an understanding of the meaning of the value, the meaning of its absence, and the way in which its absence (or confrontation) is promoted in an ironic or subversive manner. These results suggest that LLMs fall short in processing conceptual complexity and social opposition. However, these observations highlight the advantage of the 2-stage pipeline: by combining multimodal LLMs in a divide-and-conquer manner, it mitigates some of their shortcomings, achieving more robust performance both at the \emph{value} level and when aggregating per movie.}


\revision{\paragraph{Continuous Pretraining} Continuous pretraining did not improve any of the models we used. We attribute this to the combination of two main factors: (i) The relatively small dataset used for the continuous pretraining ($<$1000 videos), (ii) The large number of unbalanced classes, as presented in Figure \ref{fig:value_counts}. Future work should explore the impact of each of these factors. }

\paragraph{Limitations and Discussion} This study is not without limitations. \revision{First, we have sampled 91 influencers and 885 videos from TikTok. Our manual annotation process, included two independent expert annotators, and a third consolidator. This process ensured quality, yet limited the size of the dataset and the generalization beyond the TikTok platform. While this work explores the different pipelines and methods for value extraction, future work will incorporate a larger dataset, allowing for better generalization even for values infrequent in the current dataset.} 

Second, we have sampled influencers from seven specific genres. In our dataset, we found much variability across the genres in content and style. The genres we explored were identified as the most important for children and adolescents. TikTok, however, is extremely diverse and other genres may become prominent. Thus, future studies must verify that our results generalize across TikTok. 

A third issue stems from the movie-to-script conversion. The quality of the generated scripts was established in a qualitative manner by two of the authors.  However, even state-of-the-art models may fail to capture important details. For example, a humerus nonsensical TikTok showed the influencer air-driving a car in a charade game, appearing to be practically driving in the air, floating around the room  and bumping into objects. The generated script failed to reflect the fact that the influencer was depicted as actually moving in the air.

Forth, we applied the Schwartz personal values theory to define and identify values. This theory was tested across cultures and in hundreds of samples across the world \cite{Sagiv2022}, and is considered the leading theoretical framework of values in social science. At the same time, the theory was not constructed specifically within the context of social media, and does not focus on unique aspects of this context. Specialized theories can add information for future investigation \cite{shifman2025expression}. 

Finally, while this work explored the use of an array of strong models -- new models, announced periodically, may offer a different perspective (e.g., significant improvement of the direct pipeline).
\section{Conclusions and Implications}
\label{sec:discussion}

We have explored two pipelines for the extraction of personal values from TikTok -- a direct approach using a SOTA multimodal LLMs and an indirect, two-steps, approach using either multimodal LLM for both stages or an LLM in the first step and a supervised MLM in the second step. 
We have demonstrated that the two-steps approach mitigates some of the shortcomings of SOTA LLMs -- particularly when it comes to the processing of nuanced societal commentary and values. We further analyze the performance over individual values and over endorsed and confronted values. 

The results reported in this paper outperform state-of-the-art  results \cite{schroter2023adam,legkas2024hierocles} achieved in less challenging settings in which (i) text, rather than video, was the original medium of the input, and (ii) the number of examples available for training was an order of magnitude larger.\footnote{A direct comparison is unattainable due to the very different nature of the data in terms of source, medium and style.}

The methods presented have important implications for research and society. TikTok emerged as a leading platform among youth \cite{anderson2023teens, woodward2024}, yet at the same time produces worry among parents, educators, and decision makers \cite{conte2024scrolling}. By applying value detection, researchers can study the prevalence  transition, and effect of different values in TikTok,  at scale. Such research can provide society with much needed information regarding the social environment of young people. Importantly, rather than approaching social media influence as a monolithic concern, parents can use our value identification framework to have more nuanced conversations with their children about the content they consume. 
In addition, the methods presented here have important implications for AI alignment, platform governance, and design choices. 

It is important to note that advertisers and influencers can make similar use of the method, in order to target youth more effectively. We follow \citet{deVeirman2019influencer} in call for policy makers and educators to mitigate misuse of online platforms by enabling youth to better identify patterns of influence in social media.

\revision{
\section{Ethical Considerations}
\label{sec:ethics}
Social media platforms create a semi-public sphere where influencers actively post content not just for their friends, but for a wider audience. Influencers often use their online presence as their main source of income, putting vast effort to maximize exposure and engagement \cite{carpiano2023confronting,diresta2024invisible}. Therefore, studying influencer culture and ecosystem is not only `fair-practice' under the unobtrusive observer paradigm \cite{kozinets2020}, but also crucial for understanding how values are being monetized by influencers targeting children.  
We note that eight accounts in our data feature minors, three of these are parents accounts. The legal guardians are active participants and supporters of the other five accounts. }

\revision{The dataset curated for this work is publicly available and collected only via public APIs with no violation of TikTok terms of service. The annotated dataset available at [URL] contains only the TikTok IDs and the annotations, in order to allow influencers maintaining privacy should they choose to delete their content.}

\revision{The societal need to understand the ecosystem that targets and influences minors balances and mitigates these ethical concerns. Moreover, we believe that only such understanding will allow parents, educators, and policy makers, to support minors -- a vulnerable population --  in facing potential risks resulting from the broad use of social media. 
}

\bibliography{values_JAIR}
\newpage
\appendix


\section{Prompts and Responses}
\label{app:prompts}
In this appendix we provide the prompts used for value extraction (direct and 2-step extraction) and the prompt used in converting a movie to textual script, along with an example of a generated script and a the values extracted from it. 

\subsection{Prompt for Value Extraction}
\label{app:subsec:values_prompt}
\begin{figure*}
    \centering
    \includegraphics[width=1\linewidth]{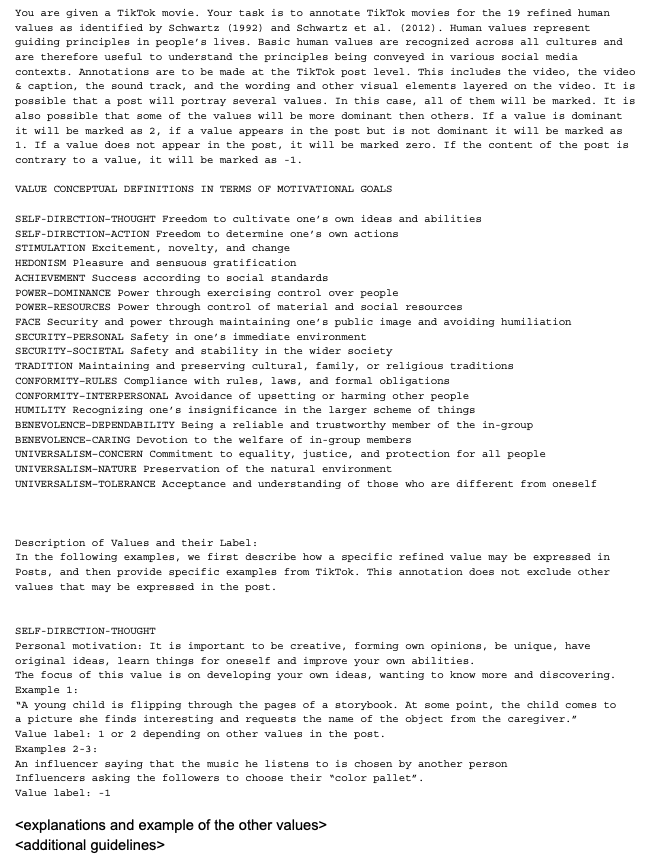}
    \caption{Annotation guidelines and prompt for direct extraction.}
    \label{fig:direct_prompt}
\end{figure*}

\begin{figure*}
    \centering
    \includegraphics[width=1\linewidth]{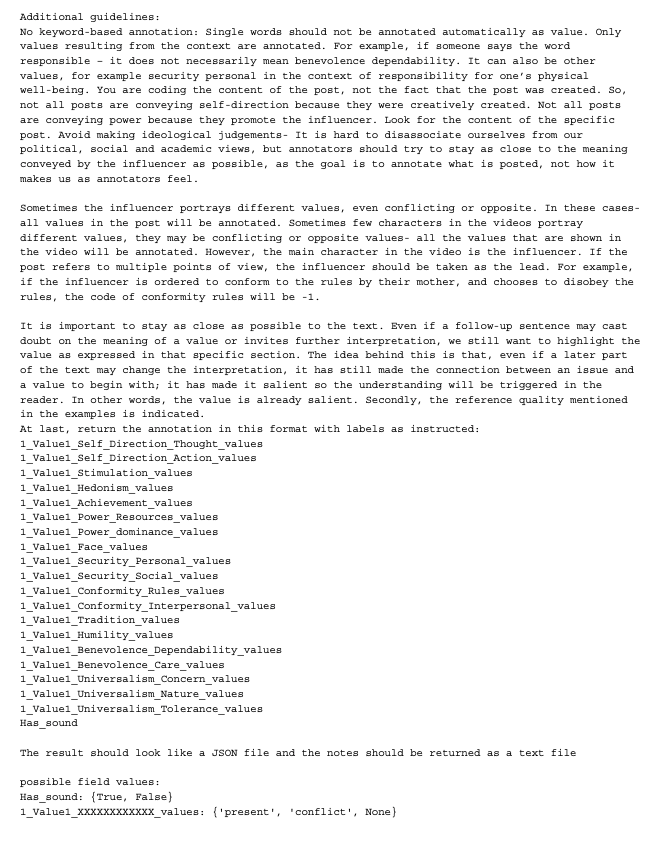}
    \caption{Additional guidelines, appended after all the examples (code book).}
    \label{fig:additional_guidlines}
\end{figure*}

The prompt used for direct extraction is provided at Figures \ref{fig:direct_prompt} and \ref{fig:additional_guidlines}. This very 
same ``prompt'' served as the annotation guidelines (code book) offered to annotators. Note that the figure contains an example (``few shot'') for only one value \texttt{Self-direction Thought} while the actual prompt (and the annotation guidelines) provided examples for each of the values.
The same prompt was used for few-shot extraction from scripts in the 2-step extraction, the sole difference is that the opening line is ``You are given a \emph{movie script} of a TikTok video'', rather than ``You are given a TikTok video''. The full code book (prompt) will be available at [URL to be added in camera ready version].

\subsection{Prompt for Script Extraction}
\label{app:subsec:script_prompt}

\begin{figure*}
    \centering
    \includegraphics[width=1\linewidth]{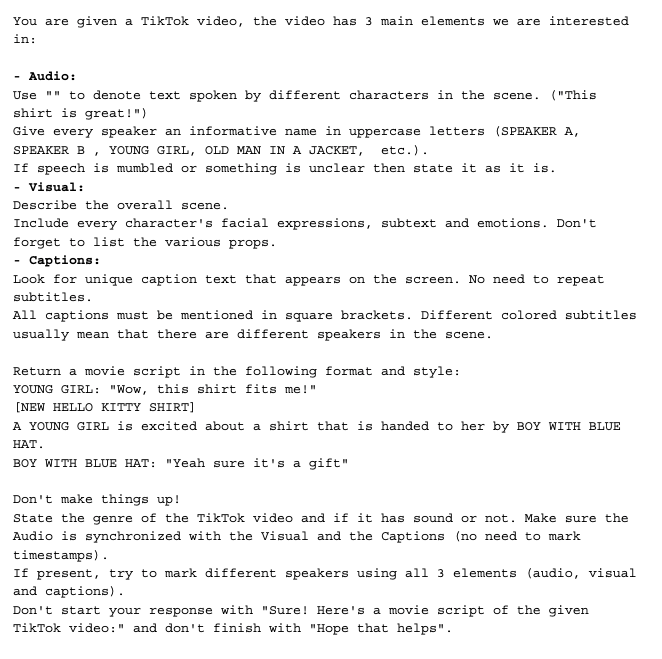}
    \caption{The prompt used to convert TikTok movies to textual scripts. }
    \label{fig:movie2script}
\end{figure*}

The prompt used to convert TikTok movies to textual scripts is available in Figure \ref{fig:movie2script}; the script generated by applying this prompt on TikTok ID 7341895431883967746 is presented in Table \ref{tab:tennis_script}.  The manual annotations for this specific movie are: \texttt{Achievement} Endorsed Value (1), \texttt{Face} Confront value (-1). 

\section{Detailed Results}
\label{app:detailed_results}
Detailed results per value are provided in Tables \ref{tab:Endorsed_Appendix} and \ref{tab:Confronted1_Appendix}.


\section{Experimental Design: configurations and hyper-parameters} 
In this section we define the hyper-parameters used in this paper. 

\subsection{MLM Training Configuration}
\label{app:subsec:mlm_hyperparams}
All MLMs were trained using the AdamW optimizer with a learning rate of 2e-5, following standard practices for transformer fine-tuning. We employed a linear learning rate schedule with warm-up, where the number of warm-up steps was set to 10\% of the total training steps, allowing the model to gradually adapt to the task-specific data distribution and preventing early training instability.

We used 5 epochs with early stopping with patience of 2 epochs.

Throughout training, transformer backbone weights remained unfrozen, enabling full model adaptation to the nuances of human values classification rather than relying solely on pre-trained representations. 

\revision{\subsection{Multimodal Extraction using VideoMAE}
\label{app:subsec:videoMAE}  Visual embeddings were extracted by sampling 16 uniformly distributed frames from each video using PyAV and processed through VideoMAE-Base to obtain visual representations. 
}

\revision{Audio embeddings were generated by extracting 10-second audio segments using librosa \cite{librosa} and processing through Wav2Vec2-Base (facebook/wav2vec2-base) with mean pooling over the temporal dimension to produce 768-dimensional feature vectors.
}

\revision{Text embeddings were obtained by performing Optical Character Recognition (OCR) using Tesseract on three key frames (first, middle, and last) to extract visual text overlays, concatenating the extracted text, and encoding it using the SentenceTransformer model 'all-MiniLM-L6-v2' to generate 384-dimensional semantic representations. The embeddings of the three modalities were then fused by concatenation to form comprehensive multimodal representations with a total dimensionality of 768 (VideoMAE) + 768 (Wav2Vec2) + 384 (SentenceTransformer) = 1,920 features per video.
}

\newpage
\begin{sidewaystable*}
\centering

\addtolength{\tabcolsep}{-3pt} 

\scriptsize
\begin{tabular}{l|*{3}{c}|*{3}{c}|*{3}{c}|*{3}{c}|*{3}{c}|*{3}{c}|*{3}{c}|*{3}{c}|*{3}{c}}
\toprule
& \multicolumn{3}{c|}{\textbf{Gemini$^{2s}$}} & \multicolumn{3}{c|}{\textbf{Gemini$^d$}} & \multicolumn{3}{c|}{\textbf{Mistral$^{2s}$}} & \multicolumn{3}{c|}{\textbf{roBERTa$^{2s}$}}& \multicolumn{3}{c|}{\textbf{DeBERTa$^{2s}$}} & \multicolumn{3}{c|}{\textbf{roBERTa$^{2s}_{PT}$}} & \multicolumn{3}{c|}{\textbf{DeBERTa$^{2s}_{PT}$}} & \multicolumn{3}{c|}{\textbf{Longformer$^{2s}$}} & \multicolumn{3}{c}{\textbf{F-VideoMAE$^{2s}$}}\\

\textbf{Value} & P & R & F1 & P & R & F1 & P & R & F1 & P & R & F1 & P & R & F1 & P & R & F1 & P & R & F1 & P & R & F1 & P & R & F1 \\
\midrule
\textbf{Self-Direction Thought} & 0.49 & 0.33 & 0.40 & 0.47 & 0.36 & 0.40 & 0.32 & 0.61 & 0.42 & 0.43 & 0.46 & 0.44& 0.45 & 0.47 & 0.46 & 0.28 & 0.35 & 0.31 & 0.39 & 0.33 & 0.36 & 0.23 & 1.00 & 0.37 & 0.12 & 0.15 & 0.13  \\
\textbf{Self-Direction Action} & 0.43 & 0.67 & 0.53 & 0.49 & 0.56 & 0.52 & 0.34 & 0.75 & 0.47 & 0.49 & 0.45 & 0.47  & 0.37 & 0.41 & 0.38 & 0.27 & 0.30 & 0.28 & 0.40 & 0.44 & 0.41 & 0.26 & 1.00 & 0.41 & 0.31 & 0.21 & 0.22  \\
\textbf{Stimulation} & 0.34 & 0.83 & 0.48 & 0.36 & 0.78 & 0.49 & 0.33 & 0.70 & 0.45 & 0.42 & 0.36 & 0.38 & 0.34 & 0.33 & 0.34 & 0.29 & 0.35 & 0.31 & 0.33 & 0.35 & 0.34 & 0.25 & 1.00 & 0.40 & 0.10 & 0.19 & 0.13 \\
\textbf{Hedonism} & 0.23 & 0.93 & 0.37 & 0.27 & 0.87 & 0.41 & 0.26 & 0.80 & 0.39 & 0.31 & 0.45 & 0.37 & 0.25 & 0.38 & 0.30 & 0.17 & 0.32 & 0.22 & 0.17 & 0.31 & 0.22 & 0.17 & 1.00 & 0.29 & 0.25 & 0.19 & 0.21  \\
\textbf{Achievement} & 0.61 & 0.77 & 0.68 & 0.68 & 0.71 & 0.69 & 0.57 & 0.62 & 0.59 & 0.70 & 0.56 & 0.62 & 0.49 & 0.46 & 0.47 & 0.33 & 0.29 & 0.31 & 0.47 & 0.40 & 0.43 & 0.36 & 1.00 & 0.53 & 0.44 & 0.47 & 0.46\\
\textbf{Power Dominance} & 0.46 & 0.47 & 0.46 & 0.51 & 0.36 & 0.42 & 0.39 & 0.27 & 0.32 & 0.34 & 0.44 & 0.37 & 0.25 & 0.38 & 0.30 & 0.25 & 0.38 & 0.30 & 0.24 & 0.35 & 0.28 & 0.19 & 0.99 & 0.32 & 0.23 & 0.24 & 0.23 \\
\textbf{Power Resources} & 0.69 & 0.43 & 0.53 & 0.68 & 0.32 & 0.44 & 0.47 & 0.28 & 0.35 & 0.29 & 0.36 & 0.32 & 0.29 & 0.35 & 0.31 & 0.23 & 0.32 & 0.26 & 0.29 & 0.33 & 0.31 & 0.20 & 1.00 & 0.33 & 0.12 & 0.40 & 0.17  \\
\textbf{Face} & 0.16 & 0.52 & 0.24 & 0.16 & 0.25 & 0.19 & 0.11 & 0.22 & 0.14 & 0.16 & 0.48 & 0.24 & 0.09 & 0.25 & 0.13 & 0.09 & 0.27 & 0.13 & 0.09 & 0.21 & 0.12 & 0.08 & 1.00 & 0.15 & 0.00 & 0.00 & 0.00  \\
\textbf{Security Personal} & 0.35 & 0.83 & 0.50 & 0.39 & 0.87 & 0.54 & 0.35 & 0.67 & 0.46 & 0.53 & 0.53 & 0.48 & 0.54 & 0.59 & 0.52 & 0.11 & 0.34 & 0.16 & 0.10 & 0.27 & 0.15 & 0.10 & 1.00 & 0.17 & 0.15 & 0.37 & 0.21\\
\textbf{Security Social}$^*$ & 0.00 & 0.00 & 0.00 & 0.00 & 0.00 & 0.00 & 0.00 & 0.00 & 0.00 & NA & NA & NA & NA & NA & NA & NA & NA & NA & NA & NA & NA & NA & NA & NA & NA & NA & NA\\
\textbf{Tradition}$^*$ & 0.35 & 0.55 & 0.43 & 0.29 & 0.45 & 0.35 & 0.20 & 0.39 & 0.27 & NA & NA & NA & NA & NA & NA & NA & NA & NA & NA & NA & NA & NA & NA & NA & NA & NA & NA \\
\textbf{Conformity Rules}$^*$ & 0.15 & 0.46 & 0.23 & 0.12 & 0.38 & 0.18 & 0.11 & 0.46 & 0.18 & NA & NA & NA & NA & NA & NA & NA & NA & NA & NA & NA & NA & NA & NA & NA & NA & NA & NA\\
\textbf{Conformity Interpersonal} & 0.40 & 0.20 & 0.27 & 0.40 & 0.20 & 0.26 & 0.10 & 0.12 & 0.10 & 0.19 & 0.45 & 0.26 & 0.15 & 0.38 & 0.21 & 0.13 & 0.38 & 0.19 & 0.10 & 0.27 & 0.15 & 0.10 & 1.00 & 0.18 & 0.10 & 0.78 & 0.18\\
\textbf{Humility} & 0.35 & 0.14 & 0.21 & 0.00 & 0.00 & 0.00 & 0.19 & 0.14 & 0.16 & 0.10 & 0.43 & 0.17& 0.10 & 0.46 & 0.16 & 0.06 & 0.32 & 0.10 & 0.11 & 0.49 & 0.18 & 0.07 & 1.00 & 0.13 & 0.07 & 0.09 & 0.08\\
\textbf{Benevolence Care} & 0.63 & 0.486 & 0.53 & 0.50 & 0.42 & 0.46 & 0.53 & 0.37 & 0.43 & 0.30 & 0.39 & 0.34 & 0.29 & 0.36 & 0.32 & 0.26 & 0.33 & 0.29 & 0.23 & 0.28 & 0.25 & 0.21 & 1.00 & 0.35 & 0.23 & 0.48 & 0.26 \\
\textbf{Benevolence Dependability} & 0.43 & 0.39 & 0.41 & 0.46 & 0.34 & 0.38 & 0.34 & 0.35 & 0.35 & 0.55 & 0.45 & 0.48 & 0.57 & 0.38 & 0.42 & 0.14 & 0.24 & 0.17 & 0.23 & 0.37 & 0.28 & 0.16 & 1.00 & 0.28 & 0.24 & 0.65 & 0.32 \\
\textbf{Universalism Concern}$^*$ & 0.33 & 1.00 & 0.50 & 0.25 & 1.00 & 0.40 & 0.03 & 1.00 & 0.05 & NA & NA & NA & NA & NA & NA & NA & NA & NA & NA & NA & NA & NA & NA & NA & NA & NA & NA \\
\textbf{Universalism Nature}$^*$ & 0.67 & 0.22 & 0.33 & 0.50 & 0.11 & 0.18 & 0.10 & 0.33 & 0.16 & NA & NA & NA & NA & NA & NA & NA & NA & NA & NA & NA & NA & NA & NA & NA & NA & NA & NA\\
\textbf{Universalism Tolerance}$^*$ & 0.33 & 0.20 & 0.25 & 0.20 & 0.20 & 0.20 & 0.05 & 0.40 & 0.10 & NA & NA & NA & NA & NA & NA & NA & NA & NA & NA & NA & NA & NA & NA & NA & NA & NA & NA \\
\bottomrule
\end{tabular}
\caption{\revision{Endorsed Values: Precision, Recall and F-scores for each model and value. Values marked with $^*$ indicate categories with fewer than 5\% of instances in the dataset. P = Precision, R = Recall, F1 = F1-score $PT$: continuous pre-trained model; $2s$ indicates 2-step extraction (extraction from scripts); $d$ indicates direct extraction from video.}}
\label{tab:Endorsed_Appendix}
\end{sidewaystable*}

\newpage


\begin{sidewaystable*}
\centering

\addtolength{\tabcolsep}{-3pt} 

\scriptsize
\begin{tabular}{l|*{3}{c}|*{3}{c}|*{3}{c}|*{3}{c}|*{3}{c}|*{3}{c}|*{3}{c}|*{3}{c}|*{3}{c}}
\toprule
& \multicolumn{3}{c|}{\textbf{Gemini$^{2s}$}} & \multicolumn{3}{c|}{\textbf{Gemini$^d$}} & \multicolumn{3}{c|}{\textbf{Mistral$^{2s}$}} & \multicolumn{3}{c|}{\textbf{roBERTa$^{2s}$}} & \multicolumn{3}{c|}{\textbf{DeBERTa$^{2s}$}} & \multicolumn{3}{c|}{\textbf{roBERTa$^{2s}_{PT}$}} & \multicolumn{3}{c|}{\textbf{DeBERTa$^{2s}_{PT}$}} & \multicolumn{3}{c|}{\textbf{Longformer$^{2s}$}} & \multicolumn{3}{c}{\textbf{F-VideoMAE}}\\
\textbf{Value} & P & R & F1 & P & R & F1 & P & R & F1 & P & R & F1 & P & R & F1 & P & R & F1 & P & R & F1 & P & R & F1 & P & R & F1 \\
\midrule
\textbf{Self-Direction Thought Neg}$^*$ & 0.09 & 0.08 & 0.09 & 0.00 & 0.00 & 0.00 & 0.02 & 0.17 & 0.04 & NA & NA & NA & NA & NA & NA & NA & NA & NA & NA & NA & NA & NA & NA & NA & NA & NA & NA \\
\textbf{Self-Direction Action Neg}$^*$ & 0.19 & 0.25 & 0.22 & 0.20 & 0.12 & 0.15 & 0.11 & 0.29 & 0.16 & NA & NA & NA & NA & NA & NA & NA & NA & NA & NA & NA & NA & NA & NA & NA & NA & NA & NA \\
\textbf{Stimulation Neg}$^*$ & 0.00 & 0.00 & 0.00 & 0.00 & 0.00 & 0.00 & 0.01 & 0.33 & 0.02 & NA & NA & NA & NA & NA & NA & NA & NA & NA & NA & NA & NA & NA & NA & NA & NA & NA & NA \\
\textbf{Hedonism Neg}$^*$ & 0.08 & 0.50 & 0.13 & 0.06 & 0.17 & 0.09 & 0.04 & 0.67 & 0.07 & NA & NA & NA & NA & NA & NA & NA & NA & NA & NA & NA & NA & NA & NA & NA & NA & NA & NA\\
\textbf{Achievement Neg}$^*$ & 0.30 & 0.68 & 0.42 & 0.38 & 0.42 & 0.40 & 0.04 & 0.26 & 0.07 & NA & NA & NA & NA & NA & NA & NA & NA & NA & NA & NA & NA & NA & NA & NA & NA & NA & NA \\
\textbf{Power Dominance Neg}$^*$ & 0.30 & 0.22 & 0.26 & 0.31 & 0.13 & 0.19 & 0.05 & 0.13 & 0.07 & NA & NA & NA & NA & NA & NA & NA & NA & NA & NA & NA & NA & NA & NA & NA & NA & NA & NA\\
\textbf{Power Resources Neg}$^*$ & 0.27 & 0.33 & 0.30 & 0.44 & 0.44 & 0.44 & 0.01 & 0.11 & 0.01 & NA & NA & NA & NA & NA & NA & NA & NA & NA & NA & NA & NA & NA & NA & NA & NA & NA & NA \\
\textbf{Face Neg} & 0.41 & 0.40 & 0.41 & 0.40 & 0.22 & 0.29 & 0.20 & 0.34 & 0.25 & 0.20 & 0.50 & 0.28 & 0.16 & 0.4 & 0.23 & 0.09 & 0.29 & 0.13 & 0.13 & 0.33 & 0.18 & 0.10 & 1.00 & 0.18 & 0.00 & 0.00 & 0.00\\
\textbf{Security Personal Neg} & 0.39 & 0.53 & 0.45 & 0.40 & 0.46 & 0.43 & 0.22 & 0.39 & 0.28 & 0.18 & 0.55 & 0.27  & 0.20 & 0.57 & 0.30 & 0.08 & 0.29 & 0.13 & 0.10 & 0.32 & 0.15 & 0.10 & 1.00 & 0.18 & 0.09 & 0.07 & 0.08\\
\textbf{Security Social Neg}$^*$ & 0.20 & 0.33 & 0.25 & 0.50 & 0.44 & 0.47 & 0.03 & 0.33 & 0.05 & NA & NA & NA & NA & NA & NA & NA & NA & NA & NA & NA & NA & NA & NA & NA & NA & NA & NA\\
\textbf{Tradition Neg}$^*$ & 0.00 & 0.00 & 0.00 & 0.50 & 0.25 & 0.33 & 0.00 & 0.00 & 0.00 & NA & NA & NA & NA & NA & NA & NA & NA & NA & NA & NA & NA & NA & NA & NA & NA & NA & NA\\
\textbf{Conformity Rules Neg}$^*$ & 0.07 & 0.38 & 0.11 & 0.10 & 0.46 & 0.17 & 0.04 & 0.38 & 0.07 & NA & NA & NA & NA & NA & NA & NA & NA & NA & NA & NA & NA & NA & NA & NA & NA & NA & NA\\
\textbf{Conformity Interpersonal Neg} & 0.43 & 0.27 & 0.33 & 0.56 & 0.18 & 0.27 & 0.32 & 0.34 & 0.33 & 0.30 & 0.52 & 0.38 & 0.27 & 0.47 & 0.34 & 0.16 & 0.29 & 0.20 & 0.16 & 0.28 & 0.20 & 0.13 & 1.00 & 0.23 & 0.18 & 0.76 & 0.25 \\
\textbf{Humility Neg} & 0.43 & 0.32 & 0.37 & 0.58 & 0.15 & 0.24 & 0.27 & 0.20 & 0.23 & 0.52 & 0.45 & 0.48 & 0.42 & 0.41 & 0.41 & 0.34 & 0.36 & 0.35 & 0.33 & 0.34 & 0.33 & 0.24 & 1.00 & 0.39 & 0.38 & 0.44 & 0.41\\
\textbf{Benevolence Care Neg} & 0.68 & 0.17 & 0.27 & 0.56 & 0.08 & 0.14 & 0.22 & 0.21 & 0.22 & 0.26 & 0.56 & 0.35  & 0.26 & 0.46 & 0.33 & 0.17 & 0.37 & 0.23 & 0.17 & 0.34 & 0.23 & 0.00 & 0.00 & 0.00 & 0.34 & 0.07 & 0.09\\
\textbf{Benevolence Dependability Neg}$^*$ & 0.27 & 0.24 & 0.26 & 0.42 & 0.12 & 0.19 & 0.04 & 0.10 & 0.06 & 0.26 & 0.56 & 0.35  & 0.00 & 0.00 & 0.00 & 0.00 & 0.00 & 0.00 & 0.00 & 0.00 & 0.00 & 0.00 & 0.00 & 0.00 & 0.00 & 0.00 & 0.00    \\
\textbf{Universalism Concern Neg}$^*$ & 0.00 & 0.00 & 0.00 & 0.00 & 0.00 & 0.00 & 0.00 & 0.00 & 0.00 & NA & NA & NA & NA & NA & NA & NA & NA & NA & NA & NA & NA & NA & NA & NA & NA & NA & NA\\
\textbf{Universalism Nature Neg}$^*$ & 0.00 & 0.00 & 0.00 & 0.00 & 0.00 & 0.00 & 0.00 & 0.00 & 0.00 & NA & NA & NA & NA & NA & NA & NA & NA & NA & NA & NA & NA & NA & NA & NA & NA & NA & NA\\
\textbf{Universalism Tolerance Neg}$^*$ & 0.10 & 1.00 & 0.18 & 0.10 & 1.00 & 0.18 & 0.01 & 1.00 & 0.02 & NA & NA & NA & NA & NA & NA & NA & NA & NA & NA & NA & NA & NA & NA & NA & NA & NA & NA\\
\bottomrule
\end{tabular}
\caption{\revision{Confronted Values: Precision, Recall and F-scores for each model and value. P = Precision, R = Recall, F1 = F1-score $PT$: continuous pre-trained model; $2s$ indicates 2-step extraction (extraction from scripts); $d$ indicates direct extraction from video.}}
\label{tab:Confronted1_Appendix}
\end{sidewaystable*}
\end{document}